\documentclass[11pt,a4paper]{article}
\usepackage[hyperref]{emnlp2020}
\usepackage{times}
\usepackage{latexsym}
\usepackage{amsmath}
\usepackage{forest}

\usepackage{microtype}

\aclfinalcopy 

\title{Incorporate Semantic Structures into Machine Translation Evaluation via UCCA}

\author{Jin Xu$^{1}$, Yinuo Guo$^{2}$, Junfeng Hu$^{2}$\\
  $^{1}$Yuanpei College, Peking University\\
  $^{2}$Key Laboratory of Computational Linguistics, School of EECS, Peking University\\
  \texttt{\{jinxu, gyn0806, hujf\}@pku.edu.cn}\\
  }

\date{}

\begin{document}
\maketitle
\begin{abstract}
\textit{Copying mechanism} has been commonly used in neural paraphrasing networks and other text generation tasks, in which some important words in the input sequence are preserved in the output sequence. 
Similarly, in machine translation, we notice that there are certain words or phrases appearing in all good translations of one source text, and these words tend to convey important semantic information. 
Therefore, in this work, we define words carrying important semantic meanings in sentences as \textit{semantic core words}.
Moreover, we propose an MT evaluation approach named \textit{Semantically Weighted Sentence Similarity (SWSS)}. It leverages the power of UCCA to identify semantic core words, and then calculates sentence similarity scores on the overlap of semantic core words.
Experimental results show that SWSS can consistently improve the performance of popular MT evaluation metrics which are based on lexical similarity.
\end{abstract}

\section{Introduction}

Machine Translation Evaluation (MTE) is to evaluate the quality of sentences produced by Machine Translation (MT) systems. 
Most automatic MT evaluation metrics compare the candidate sentences from MT systems with reference sentences from human translation to produce a similarity score, in contrast some other reference-less metrics directly compare candidate sentences and source sentences.

According to the observation of well-translated sentences, we find out that there are certain words or phrases appearing in all good translations of one source text. This phenomenon is consistent with the intuition of copying mechanism~\cite{copy}, which has been widely used in lots of text generation tasks. In the field of MT evaluation, Meteor++~\citep{meteor++} firstly proposes the concept of \textit{copy knowledge} to define the words with copy property, and it further incorporates the copy knowledge into Meteor~\citep{meteor} to improve its performance. Specifically, it attempts to find copy words of references and candidate sentences, and uses the overlap of these words to modify the calculation of precision and recall of Meteor. However, Meteor++ uses named entities as an alternative to copy knowledge in its experiments, resulting in a limited range of selected copy words and a slight improvement. 

\begin{figure}[tbp!]
\centering
\includegraphics[width=7.5cm]{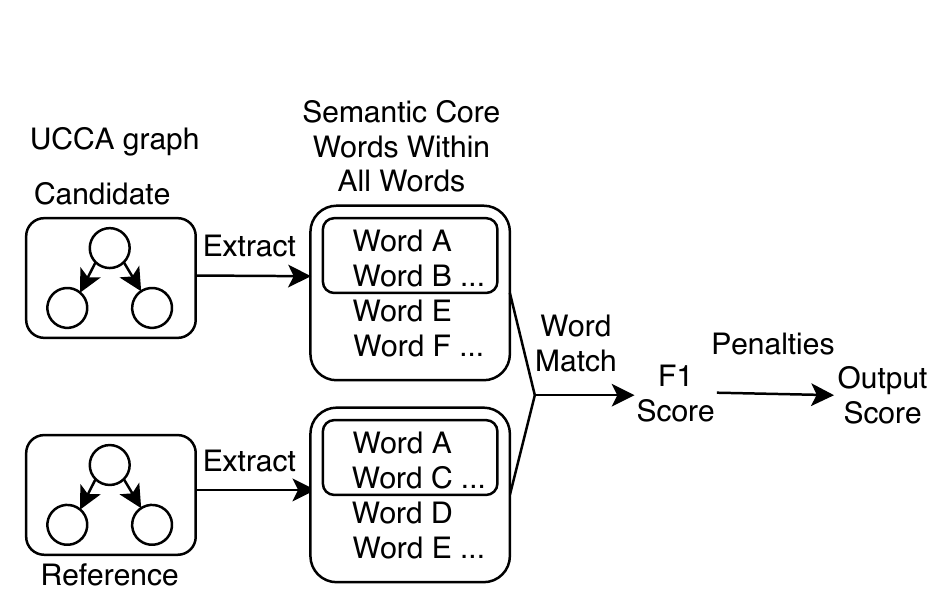}
\caption{An illustration of the process of SWSS.}
\label{fig:0}
\end{figure}

In this work, we argue that words undertaking important semantic meanings should be exactly expressed during the translation procedure, which we define as semantic core words.
This concept is much more general and closer to linguistic intuition compared to the copy knowledge used in Meteor++. In order to apply semantic core words in the process of MT evaluation, we design a mechanism named \textit{Semantically Weighted Sentence Similarity (SWSS)} illustrated in Figure \ref{fig:0}. Firstly, SWSS extracts semantic core words according to the annotated semantic labels in Universal Conceptual Cognitive Annotation (UCCA)~\citep{UCCA}, a multi-layered semantic representation. UCCA is an appealing candidate for this mechanism as it includes a lot of fundamental semantic
phenomena, such as verbal, nominal and adjectival
argument structures and their inter-relations. Also, semantic units in UCCA are anchored in the text, which simplifies the aligning procedure a lot.
With the assumption that all high-quality translations should have the same semantic core words, SWSS then calculates precision and recall based on the overlap of semantic core words between sentence pairs and their corresponding F1 scores.
Finally, we modify the F1 score according to the differences of two UCCA representations. For example, {\it Scenes} are involved in the penalties, which are essential nodes in UCCA indicating actions and states of the sentences.
Our experimental results show that SWSS can be combined with other popular MT evaluation metrics to improve their performance significantly.
\section{Related Work}

\subsection{Machine Translation Evaluation}

BLEU ~\citep{bleu} and Meteor are two most popular MT evaluation metrics. BLEU measures n-grams overlapping between the candidate sentences and reference sentences, while Meteor aligns words and phrases to calculate a modified weighted F-score. The two metrics are based on lexical similarity but somehow neglect semantic structure information of the sentences.
	
Efforts have been made to incorporate linguistic features and resources into MT evaluation. RED~\citep{red} makes use of dependency tree and MEANT~\citep{meant} makes use of semantic parser. Categories such as part-of-speech~\citep{mt18} and named entity~\citep{wmt12} also have their effects. In order to complement WordNet~\citep{wordnet} and paraphrase table in Meteor, Meteor++2.0 ~\citep{meteor++2} applies syntactic-level paraphrase knowledge.

\subsection{Semantic Representation}


Semantic representation focuses on how meaning is expressed in a sentence. 
Some semantic representation frameworks such as UNL~\citep{unl} and AMR~\citep{amr} use concept nodes to represent content words of sentence, and use directed edges with labels to indicate the semantic relation between nodes. 

\begin{figure}[tb!]
\centering
\small
\begin{forest}
 [,fill=black, circle,
 [,edge=->, edge label={node[midway,left]{H}}, fill=black, circle,
 	[,edge=->, edge 			label={node[midway,left]{A}}, fill=black, circle, l=13mm,
		[John, edge=->, edge 			label={node[midway,left]{C}}],
		[and, edge=->, edge 			label={node[midway,left]{N}}],
		[Mary, edge=->, edge 			label={node[midway,left]{C}}] 	
 	],
 	[bought,edge=->, edge 			label={node[midway,right]{P}}, l=13mm],
 	[,edge=->, edge 			label={node[midway,right]{A}}, fill=black, circle, l=13mm,
 		[the, edge=->, edge 			label={node[midway,left]{E}}, l=13mm],
 		[sofa, edge=->, edge 			label={node[midway,left]{C}}, l=13mm, name=sofa],{\draw[<-, dashed] (sofa) to[out=east, in=west]
 node[very near start,above]{\normalsize A}
 (h2);},
 		[,edge=->, edge 			label={node[midway,right]{E}}, fill=black, circle, l=11mm, name=h2, s sep=10mm,
			[I,edge=->, edge 			label={node[midway,right]{A}}],
			[sold,edge=->, edge 			label={node[midway,right]{P}}] 		
 		]
 		],
 	[together, edge=->, edge 			label={node[midway,right]{D}},l=13mm]
]
]
\end{forest}
\caption{~~UCCA representation of sentence "John and Mary bought the sofa I sold together". Labels include \textit{Parallel Scene (H)}, \textit{Participant (A)}, \textit{Process (P)}, \textit{Adverbial (D)}, \textit{Center (C)}, \textit{Connector (N)}, \textit{Elaborator (E)}. Dash line indicates a secondary semantic relation. There are two scenes in this sentence, the whole sentence and "I sold (sofa)".}
\label{graph:1}
\end{figure}
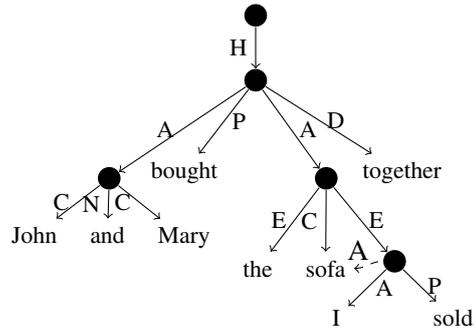

UCCA is a novel multi-layered semantic representation framework, which converts a sentence into a directed acyclic graph (DAG). Leaf nodes of UCCA graph correspond to words in the sentence, and a non-leaf node represents the combination of meanings of its child nodes. A parent node and a child node are connected by a directed edge which demonstrates the semantic role of the child node in the meaning of the parent node. Figure \ref{graph:1} is an example of UCCA representation.

Scene is an essential concept in UCCA. A scene describes some movement, action or a state in the sentence. Scene nodes in UCCA representation may be connected to the root node, or embedded in other scenes as arguments or modifiers. A scene node has a main relation, either a \textit{Process} or a \textit{State}, and may have some \textit{Participants} or some \textit{Adverbials}. These non-scene nodes may also have inner structure.

UCCA has been applied in many fields of Natural Language Processing. SAMSA~\citep{ts} is a Text Simplification evaluation metric that defines minimal center of UCCA representation and compares simplified text with the minimal centers of original sentences. It is also used in evaluation of faithfulness in Grammatical Error Correlation ~\citep{gec} and human MT evaluation ~\citep{hume}.

\section{Proposed Method}

\subsection{Semantic Core Words}

The most popular MT evaluation metrics such as BLEU and Meteor are based on lexical similarity. This kind of metrics cannot obtain insight into semantic structure of the whole sentence. Our proposed semantic core words are extracted from UCCA semantic structures and used to improve these lexical metrics as we expect them to play the role of copy words.

It is a linguistic intuition that some words carry more semantic information than other words in a sentence. For example, a modifier is usually less important than the word it modifies. In this paper, We define words that have important semantic information as semantic core words. According to their semantic importance, they are expected to be accurately translated during translation. Therefore, we assume that in all good translation results of a specific sentence, the set of semantic core words should be the same, behaving like copy words.
	
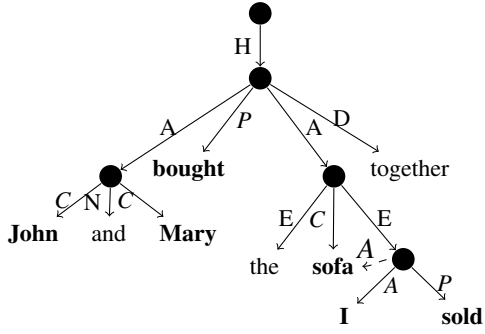
\begin{figure}[tb!]
\centering
\small
\begin{forest}
 [,fill=black, circle,
 [,edge=->, edge label={node[midway,left]{H}}, fill=black, circle,
 	[,edge=->, edge 			label={node[midway,left]{A}}, fill=black, circle, l=13mm,
		[{\bf John}, edge=->, edge 			label={node[midway,left]{\it C}}],
		[and, edge=->, edge 			label={node[midway,left]{N}}],
		[{\bf Mary}, edge=->, edge 			label={node[midway,left]{\it C}}] 	
 	],
 	[{\bf bought},edge=->, edge 			label={node[midway,right]{\it P}}, l=13mm],
 	[,edge=->, edge 			label={node[midway,right]{A}}, fill=black, circle, l=13mm,
 		[the, edge=->, edge 			label={node[midway,left]{E}}, l=13mm],
 		[{\bf sofa}, edge=->, edge 			label={node[midway,left]{\it C}}, l=13mm, name=sofa],{\draw[<-, dashed] (sofa) to[out=east, in=west]
 node[very near start,above]{\normalsize \it A}
 (h2);},
 		[,edge=->, edge 			label={node[midway,right]{E}}, fill=black, circle, l=11mm, name=h2, s sep=10mm,
			[{\bf I},edge=->, edge 			label={node[midway,right]{\it A}}],
			[{\bf sold},edge=->, edge 			label={node[midway,right]{\it P}}] 		
 		]
 		],
 	[{together}, edge=->, edge 			label={node[midway,right]{D}},l=13mm]
]
]
\end{forest}
\caption{~~An example of semantic core words. The sentence is the same with Figure \ref{graph:1}. All semantic core words are bold and the semantic labels of related edges are italic.}
\label{graph:2}
\end{figure}

We extract semantic core words of a sentence from its UCCA semantic representation. The lowest semantic role label in the representation for each word is considered, which also indicates the most basic semantic role of a word. A word whose lowest semantic role is \textit{Process}, \textit{State}, \textit{Participant} or \textit{Center} is identified as semantic core words. Figure \ref{graph:2} marks semantic core words of the example sentence. The result is consistent with our intuition of which word has important meaning in this sentence.

\subsection{Word Matching}

After semantic core words are extracted from UCCA representations, a word matching algorithm should be applied in order to match all words between the two sentences. In this paper, we use a stemming algorithm. Two words are matched if they have the same stem. 
	
We count how many semantic core words in a candidate sentence can be matched to any semantic core words in the reference sentence, and compute the proportion as precision. Similarly, we calculate the matched proportion of semantic core words in reference sentence as recall. In our word matching algorithm, it is possible that a word in a sentence is matched to multiple words in the other sentence because they all have the same word stem. However, just like what is conducted in BLEU, a word cannot be contained in multiple matching pairs. The precision and recall are then used to calculate F1 score. We use F1 score here to ensure that SWSS is symmetrical and can be directly used as a sentence similarity metric. 
\begin{equation}
\begin{aligned}
P& =\frac{\sum_i{w(h_i)\cdot m(h_i)}}{\sum_i{w(h_i)}}\\
R& =\frac{\sum_i{w(r_i)\cdot m(r_i)}}{\sum_i{w(r_i)}}\\
F_1& =\frac{2P\cdot R}{P+R}
\end{aligned}
\end{equation}
Take the calculation of precision as an example. $h_i$ means each semantic core word in the candidate sentence, and $w(h_i)$ is its weight. Though in this paper the weight is fixed to 1, it can be fine-tuned or trained in future work. If $h_i$ can be matched to any semantic core word in the reference sentence, $m(h_i)$ is set to 1, otherwise $m(h_i)$ is set to 0. However, $m(h_i)$ can also be different values related to matching type like the operation in Meteor, which might be conducted in future work.
	
A fact is that the UCCA parser we used might occasionally produce an analysis result in which there are no semantic core words in a sentence, which causes division by zero during calculation. In these cases a fixed score $\omega$ is used as an alternative.

\begin{table*}[htb!]
\centering
\small
\begin{tabular}{|c|c|c||c|c||c|c|}
\hline
Base Model & \multicolumn{2}{c||}{BLEU} & \multicolumn{2}{c||}{Meteor}     & \multicolumn{2}{c|}{Meteor++}   \\ \hline
Method     & None    & +UCCA           & None         & +UCCA          & None       & +UCCA          \\ \hline
\multicolumn{7}{|c|}{WMT15}                                                                                \\ \hline
cs-en      & 0.377   & \textbf{0.418}  & 0.605          & \textbf{0.609} & 0.610          & \textbf{0.613} \\ \hline
de-en      & 0.420   & \textbf{0.464}  & 0.620          & \textbf{0.638} & 0.637          & \textbf{0.651} \\ \hline
fi-en      & 0.378   & \textbf{0.444}  & 0.645          & \textbf{0.668} & 0.661          & \textbf{0.679} \\ \hline
ru-en      & 0.445   & \textbf{0.477}  & 0.628          & \textbf{0.634} & 0.620          & \textbf{0.629} \\ \hline
Average    & 0.405   & \textbf{0.451}  & 0.624          & \textbf{0.637} & 0.632          & \textbf{0.643} \\ \hline
\multicolumn{7}{|c|}{WMT16}                                                                                \\ \hline
cs-en      & 0.484   & \textbf{0.508}  & \textbf{0.649} & 0.646          & \textbf{0.656} & 0.651          \\ \hline
de-en      & 0.367   & \textbf{0.394}  & 0.503          & \textbf{0.520} & 0.507          & \textbf{0.523} \\ \hline
fi-en      & 0.325   & \textbf{0.368}  & 0.537          & \textbf{0.548} & 0.557          & \textbf{0.564} \\ \hline
ro-en      & 0.418   & \textbf{0.451}  & 0.626          & \textbf{0.633} & 0.625          & \textbf{0.632} \\ \hline
ru-en      & 0.377   & \textbf{0.413}  & 0.574          & \textbf{0.578} & 0.583          & \textbf{0.585} \\ \hline
tr-en      & 0.333   & \textbf{0.401}  & 0.609          & \textbf{0.638} & 0.600          & \textbf{0.628} \\ \hline
Average    & 0.384   & \textbf{0.423}  & 0.583          & \textbf{0.594} & 0.588          & \textbf{0.597} \\ \hline
\end{tabular}
\caption{~~Segment-level Pearson correlation comparison between base model and the combination of SWSS and base model. The smoothing parameter $X$ of Meteor++ is set to 8, which is used on WMT15 dataset in its paper.}
\label{tab:4}
\end{table*}

\subsection{Penalty and Combination}

According to the intuition that good translation results of a specific sentence should have similar semantic structures, we introduce three penalties concerning statistical differences of two UCCA representations.
\begin{itemize}
    \item The ratio between counts of scenes of two representations. Let $S_1$, $S_2$ be the counts of scenes, the penalty $P_S$ is $1-\min(S_1, S_2)/\max(S_1, S_2)$.
    \item The ratio between counts of nodes of two representations. Let $N_1$, $N_2$ be the counts of nodes, the penalty $P_N$ is $1-\min(N_1, N_2)/\max(N_1, N_2)$.
    \item The ratio between counts of edges towards critical semantic roles of two representations, which are \textit{Process}, \textit{State} and \textit{Participant}. This count is the sum of count of scenes and count of all arguments in the sentence. Let $E_1$, $E_2$ be the counts of these edges, the penalty $P_E$ is $1-\min(E_1, E_2)/\max(E_1, E_2)$.
\end{itemize}
The three penalties are set to 0 if the counts are equal and their upper bound is 1. Additionally, we also notice that the average word count of a sentence pair can act as another penalty $Len$. Applying the four penalties, the final score is calculated by the equation below. All parameters here are tunable.
\begin{equation}
\begin{aligned}
Score = F_1 \cdot\exp( &- \alpha_1\cdot P_S -\alpha_2\cdot P_N \\
 &- \alpha_3\cdot P_E -\alpha_4\cdot Len)
\end{aligned}
\end{equation}
The SWSS score is calculated independently. Therefore, as a semantic structure-based component, it can be further combined with other MT evaluation metrics to obtain a more accurate evaluation metric. For example, we can obtain a simple weighted model of SWSS and Meteor by tuning the weight $\beta$ below.
\begin{equation}
SWSS \star Meteor = Meteor + \beta\cdot Score
\end{equation}

\section{Experiments}

\subsection{Data}

SWSS is evaluated on WMT15~\citep{wmt15} and WMT16 metric task~\citep{wmt16} evaluation sets and is tuned on WMT17 metric task ~\citep{wmt17} evaluation set. The datasets are composed of pairs of system output sentences and reference sentences, and also corresponding human evaluation scores for the output sentences. The evaluation set of WMT15 has 4 language pairs and each has 500 sentence pairs. WMT16 dataset has 6 language pairs and WMT17 dataset has 7 language pairs, and each has 560 sentence pairs. Performance of a metric is evaluated by Pearson correlation between scores provided by the metric and the human evaluation scores.

\subsection{Settings}

The parameters of SWSS are tuned on the dataset from WMT17 metric task and are listed in Table \ref{tab:2}. We use SpaCy library\footnote{\url{https://spacy.io/}} for word tokenization. Word stems are extracted with Porter stemming algorithm~\citep{porter}. UCCA representations are parsed with the pre-trained model of TUPA~\citep{tupa}.

\subsection{Results}

\begin{table}[tb!]
\small
\centering
\begin{tabular}{|c|c|c|c|}
\hline
$\alpha_1$ & 0.2 &  $\alpha_4$ & 0.01\\
$\alpha_2$ & 1 & $\beta$ & 0.2\\
$\alpha_3$ & 0.5&  $\omega$ & 0.5\\
\hline
\end{tabular}
\caption{~~Parameters of SWSS in experiments.}
\label{tab:2}
\end{table}
	
SWSS is combined with base models including BLEU, Meteor and Meteor++. Table \ref{tab:4} shows that the combined models lead to significant improvement of Pearson correlation compared to the base models. It can be inferred that adding SWSS as a component to MT evaluation metrics based on lexical similarity can improve their performance. The results also indicates that SWSS performs better than Meteor++, as SWSS regards all semantic core words as copy words while Meteor++ uses only named entities in its experiments. Semantic core words is clearly a good and large-scale representation of copy words, according to the results.

\begin{table}[tb!]
\small
\centering
\begin{tabular}{|c|c|c|c|c|}
\hline
Method     & +UCCA          & -repr          & -len           & None           \\ \hline
\multicolumn{5}{|c|}{WMT15}                                                    \\ \hline
cs-en      & \textbf{0.609} & 0.599          & 0.606          & 0.605          \\ \hline
de-en      & 0.638          & \textbf{0.641} & 0.631          & 0.620          \\ \hline
fi-en      & \textbf{0.668} & 0.662          & 0.666          & 0.645          \\ \hline
ru-en      & \textbf{0.634} & 0.622          & \textbf{0.634} & 0.628          \\ \hline
Average    & \textbf{0.637} & 0.631          & 0.634          & 0.624          \\ \hline
\multicolumn{5}{|c|}{WMT16}                                                    \\ \hline
cs-en      & 0.646          & 0.648          & 0.645          & \textbf{0.649} \\ \hline
de-en      & \textbf{0.520} & 0.512          & 0.512          & 0.503          \\ \hline
fi-en      & \textbf{0.548} & 0.541          & 0.543          & 0.537          \\ \hline
ro-en      & \textbf{0.633} & 0.631          & 0.627          & 0.626          \\ \hline
ru-en      & 0.578          & \textbf{0.581} & 0.564          & 0.574          \\ \hline
tr-en      & \textbf{0.638} & 0.632          & 0.627          & 0.609          \\ \hline
Average    & \textbf{0.594} & 0.591          & 0.586          & 0.583          \\ \hline
\end{tabular}
\caption{~~Results of ablation experiments. "+UCCA" is the complete SWSS model combined with Meteor, "-repr" means the penalties based on UCCA representation ($P_S$, $P_N$, $P_E$) are removed, "-len" means the length penalty is removed, and "None" contains only Meteor without SWSS.}
\label{tab:5}
\end{table}

We also conduct ablation study to figure out whether the penalties we have introduced are redundant or not. The base model is the combination of SWSS and Meteor. If we remove the representation penalties or the length penalty from the base model, it can be found out from Table \ref{tab:5} that the modified models have lower correlation than the complete model. The result with $p < 0.05$ proves that these penalties have a positive effect on the mechanism.

\section{Conclusion}
In this paper, we propose Semantically Weighted Sentence Similarity (SWSS), which leverages the power of UCCA to identify semantic core words, and then calculates a similarity score for machine translation evaluation. 
Inspired by copying mechanism used in sequence generation tasks, we argue that semantic core words, which carry important meaning in the sentence, should exist in all good translations. 
Additionally, SWSS also uses penalties based on the differences between UCCA structures and sentence lengths, including the concept of Scene in UCCA, in order to make the output scores more accurate.
Experimental results show that SWSS can produce a higher correlation in MT evaluation when combined with lexical MT evaluation metrics such as BLEU and Meteor.


\bibliographystyle{acl_natbib}
\bibliography{emnlp2020}

\begin{thebibliography}{21}
\expandafter\ifx\csname natexlab\endcsname\relax\def\natexlab#1{#1}\fi

\bibitem[{Abend and Rappoport(2013)}]{UCCA}
Omri Abend and Ari Rappoport. 2013.
\newblock Universal conceptual cognitive annotation (ucca).
\newblock In \emph{Proceedings of the 51st Annual Meeting of the Association
  for Computational Linguistics (Volume 1: Long Papers)}, pages 228--238.

\bibitem[{Avramidis et~al.(2011)Avramidis, Popovic, Vilar, and
  Burchardt}]{mt18}
Eleftherios Avramidis, Maja Popovic, David Vilar, and Aljoscha Burchardt. 2011.
\newblock Evaluate with confidence estimation: Machine ranking of translation
  outputs using grammatical features.
\newblock In \emph{Proceedings of the Sixth Workshop on Statistical Machine
  Translation}, pages 65--70. Association for Computational Linguistics.

\bibitem[{Banarescu et~al.(2013)Banarescu, Bonial, Cai, Georgescu, Griffitt,
  Hermjakob, Knight, Koehn, Palmer, and Schneider}]{amr}
Laura Banarescu, Claire Bonial, Shu Cai, Madalina Georgescu, Kira Griffitt, Ulf
  Hermjakob, Kevin Knight, Philipp Koehn, Martha Palmer, and Nathan Schneider.
  2013.
\newblock Abstract meaning representation for sembanking.
\newblock In \emph{Proceedings of the 7th linguistic annotation workshop and
  interoperability with discourse}, pages 178--186.

\bibitem[{Birch et~al.(2016)Birch, Abend, Bojar, and Haddow}]{hume}
Alexandra Birch, Omri Abend, Ond{\v{r}}ej Bojar, and Barry Haddow. 2016.
\newblock \href {https://doi.org/10.18653/v1/D16-1134} {{HUME}: Human
  {UCCA}-based evaluation of machine translation}.
\newblock In \emph{Proceedings of the 2016 Conference on Empirical Methods in
  Natural Language Processing}, pages 1264--1274, Austin, Texas. Association
  for Computational Linguistics.

\bibitem[{Bojar et~al.(2017)Bojar, Graham, and Kamran}]{wmt17}
Ond{\v{r}}ej Bojar, Yvette Graham, and Amir Kamran. 2017.
\newblock Results of the {WMT}17 metrics shared task.
\newblock In \emph{Proceedings of the Second Conference on Machine
  Translation}, pages 489--513, Copenhagen, Denmark. Association for
  Computational Linguistics.

\bibitem[{Bojar et~al.(2016)Bojar, Graham, Kamran, and Stanojevi{\'c}}]{wmt16}
Ond{\v{r}}ej Bojar, Yvette Graham, Amir Kamran, and Milo{\v{s}} Stanojevi{\'c}.
  2016.
\newblock Results of the wmt16 metrics shared task.
\newblock In \emph{Proceedings of the First Conference on Machine Translation:
  Volume 2, Shared Task Papers}, pages 199--231.

\bibitem[{Buck(2012)}]{wmt12}
Christian Buck. 2012.
\newblock Black box features for the wmt 2012 quality estimation shared task.
\newblock In \emph{Proceedings of the Seventh Workshop on Statistical Machine
  Translation}, pages 91--95.

\bibitem[{Choshen and Abend(2018)}]{gec}
Leshem Choshen and Omri Abend. 2018.
\newblock \href {https://doi.org/10.18653/v1/N18-2020} {Reference-less measure
  of faithfulness for grammatical error correction}.
\newblock In \emph{Proceedings of the 2018 Conference of the North {A}merican
  Chapter of the Association for Computational Linguistics: Human Language
  Technologies, Volume 2 (Short Papers)}, pages 124--129, New Orleans,
  Louisiana. Association for Computational Linguistics.

\bibitem[{Denkowski and Lavie(2014)}]{meteor}
Michael Denkowski and Alon Lavie. 2014.
\newblock Meteor universal: Language specific translation evaluation for any
  target language.
\newblock In \emph{Proceedings of the ninth workshop on statistical machine
  translation}, pages 376--380.

\bibitem[{Gu et~al.(2016)Gu, Lu, Li, and Li}]{copy}
Jiatao Gu, Zhengdong Lu, Hang Li, and Victor~OK Li. 2016.
\newblock Incorporating copying mechanism in sequence-to-sequence learning.
\newblock \emph{arXiv preprint arXiv:1603.06393}.

\bibitem[{Guo and Hu(2019)}]{meteor++2}
Yinuo Guo and Junfeng Hu. 2019.
\newblock Meteor++ 2.0: Adopt syntactic level paraphrase knowledge into machine
  translation evaluation.
\newblock In \emph{Proceedings of the Fourth Conference on Machine Translation
  (Volume 2: Shared Task Papers, Day 1)}, pages 501--506.

\bibitem[{Guo et~al.(2018)Guo, Ruan, and Hu}]{meteor++}
Yinuo Guo, Chong Ruan, and Junfeng Hu. 2018.
\newblock Meteor++: Incorporating copy knowledge into machine translation
  evaluation.
\newblock In \emph{Proceedings of the Third Conference on Machine Translation:
  Shared Task Papers}, pages 740--745.

\bibitem[{Hershcovich et~al.(2017)Hershcovich, Abend, and Rappoport}]{tupa}
Daniel Hershcovich, Omri Abend, and Ari Rappoport. 2017.
\newblock \href {https://doi.org/10.18653/v1/P17-1104} {A transition-based
  directed acyclic graph parser for {UCCA}}.
\newblock In \emph{Proceedings of the 55th Annual Meeting of the Association
  for Computational Linguistics (Volume 1: Long Papers)}, pages 1127--1138,
  Vancouver, Canada. Association for Computational Linguistics.

\bibitem[{Lo et~al.(2012)Lo, Tumuluru, and Wu}]{meant}
Chi-kiu Lo, Anand~Karthik Tumuluru, and Dekai Wu. 2012.
\newblock Fully automatic semantic mt evaluation.
\newblock In \emph{Proceedings of the Seventh Workshop on Statistical Machine
  Translation}, pages 243--252. Association for Computational Linguistics.

\bibitem[{Miller(1998)}]{wordnet}
George~A Miller. 1998.
\newblock \emph{WordNet: An electronic lexical database}.
\newblock MIT press.

\bibitem[{Papineni et~al.(2002)Papineni, Roukos, Ward, and Zhu}]{bleu}
Kishore Papineni, Salim Roukos, Todd Ward, and Wei-Jing Zhu. 2002.
\newblock Bleu: a method for automatic evaluation of machine translation.
\newblock In \emph{Proceedings of the 40th annual meeting on association for
  computational linguistics}, pages 311--318. Association for Computational
  Linguistics.

\bibitem[{Porter et~al.(1980)}]{porter}
Martin~F Porter et~al. 1980.
\newblock An algorithm for suffix stripping.
\newblock \emph{Program}, 14(3):130--137.

\bibitem[{Stanojevi{\'c} et~al.(2015)Stanojevi{\'c}, Kamran, Koehn, and
  Bojar}]{wmt15}
Milo{\v{s}} Stanojevi{\'c}, Amir Kamran, Philipp Koehn, and Ond{\v{r}}ej Bojar.
  2015.
\newblock Results of the wmt15 metrics shared task.
\newblock In \emph{Proceedings of the Tenth Workshop on Statistical Machine
  Translation}, pages 256--273.

\bibitem[{Sulem et~al.(2018)Sulem, Abend, and Rappoport}]{ts}
Elior Sulem, Omri Abend, and Ari Rappoport. 2018.
\newblock \href {https://doi.org/10.18653/v1/N18-1063} {Semantic structural
  evaluation for text simplification}.
\newblock In \emph{Proceedings of the 2018 Conference of the North {A}merican
  Chapter of the Association for Computational Linguistics: Human Language
  Technologies, Volume 1 (Long Papers)}, pages 685--696, New Orleans,
  Louisiana. Association for Computational Linguistics.

\bibitem[{Uchida and Zhu(2001)}]{unl}
Hiroshi Uchida and Meiying Zhu. 2001.
\newblock The universal networking language beyond machine translation.
\newblock In \emph{International Symposium on Language in Cyberspace, Seoul},
  pages 26--27.

\bibitem[{Yu et~al.(2014)Yu, Wu, Xie, Jiang, Liu, and Lin}]{red}
Hui Yu, Xiaofeng Wu, Jun Xie, Wenbin Jiang, Qun Liu, and Shouxun Lin. 2014.
\newblock Red: A reference dependency based mt evaluation metric.
\newblock In \emph{Proceedings of COLING 2014, the 25th international
  conference on computational linguistics: technical papers}, pages 2042--2051.

\end{thebibliography}

\end{document}